\documentclass[11pt,a4paper]{article}
\usepackage[hyperref]{acl2018}

\usepackage{times}
\usepackage{amsmath}
\usepackage{amsthm}
\usepackage{latexsym}
\usepackage{array}
\usepackage[font={small}]{caption}
\usepackage{wrapfig}

\usepackage{multirow}
\usepackage{setspace}
\usepackage{colordvi}  
\usepackage{graphicx}
\usepackage{epsfig}
\usepackage{epstopdf}
\usepackage{relsize}
\usepackage{url}
\usepackage{breakurl}

\aclfinalcopy % Uncomment this line for the final submission
\newcommand{\com}[1]{}

\title{Simple and Effective Text Simplification Using\\
Semantic and Neural Methods}

\author{Elior Sulem, Omri Abend, Ari Rappoport \\
 Department of Computer Science, The Hebrew University of Jerusalem\\
{\tt \{eliors|oabend|arir\}@cs.huji.ac.il}
 }
\date{}

\begin{document}
\maketitle
\begin{abstract}
Sentence splitting is a major simplification operator. Here we present a simple and efficient splitting algorithm based on an automatic semantic parser. After splitting, the text is amenable for further fine-tuned simplification operations. In particular, we show that neural Machine Translation can be effectively used in this situation.
Previous application of Machine Translation for simplification suffers from a considerable disadvantage in that they are over-conservative, often failing to modify the source in any way. Splitting based on semantic parsing, as proposed here, alleviates this issue. 
Extensive automatic and human evaluation shows that the proposed method compares favorably to the state-of-the-art in combined lexical and structural simplification.

\end{abstract}

\section{Introduction} \label{sec:introduction}

Text Simplification (TS) is generally defined as the conversion of a sentence into one or more simpler sentences. 
It has been shown useful both as a preprocessing step for tasks such as Machine Translation \citep[MT;][]{M14,SP16} 
and relation extraction \citep{N16}, as well as for developing reading aids, e.g. for people with dyslexia \citep{R13} or non-native speakers \citep{S02}.

TS includes both structural and lexical operations. The main structural simplification operation is sentence splitting, namely 
rewriting a single sentence into multiple sentences while preserving its meaning.
While recent improvement in TS has been achieved by the use of neural MT (NMT) approaches \citep{Ni17,Z17,ZL17}, where TS is considered a case of monolingual translation, the sentence splitting operation has not been addressed by these systems, potentially due to the rareness of this operation in the training corpora \citep{NG14,Xu15}.

We show that the explicit integration of sentence splitting in the simplification system could also reduce conservatism, which is a grave limitation of NMT-based TS systems \citep{AM17}. Indeed, experimenting with a state-of-the-art neural system \citep{Ni17}, we find that 66$\%$ of the input sentences remain unchanged, while none of the corresponding references is identical to the source.
Human and automatic evaluation of the references (against other references), confirm that the references are indeed simpler than the source,
indicating that the observed conservatism is excessive. 
Our methods for performing sentence splitting as pre-processing allows the TS system to perform other structural (e.g. deletions) and lexical (e.g. word substitutions) operations, thus increasing both structural and lexical simplicity.

For combining linguistically informed sentence splitting with data-driven TS, two main methods have been proposed.
The first  involves hand-crafted syntactic rules, whose compilation and validation are laborious \citep{Sh14}. For example, \citet{SA14} used 111 rules for relative clauses, appositions, subordination and coordination. Moreover, syntactic splitting rules, which form a substantial part of the rules, are usually language specific, requiring the development of new rules when ported to other languages \citep[for Portuguese, French, Vietnamese, and Italian respectively]{AG10,S12,H12,BT13}.
The second method uses linguistic information for detecting potential splitting points, while splitting probabilities are learned using a parallel corpus. For example, in the system of \citet{NG14} (henceforth, {\sc Hybrid}), the state-of-the-art for joint structural and lexical TS, potential splitting points are determined by event boundaries. 

In this work, which is the first to combine structural semantics and neural methods for TS, we propose an intermediate way for performing sentence splitting, presenting Direct Semantic Splitting (DSS), a simple and efficient algorithm based on a semantic parser which supports the direct decomposition of the sentence into its main semantic constituents. 
After splitting, NMT-based simplification is performed, using the NTS system. We show that the resulting system outperforms {\sc Hybrid} in both automatic and human evaluation.

We use the UCCA scheme for semantic representation \citep{AR13}, where the semantic units are anchored in the text, 
which simplifies the splitting operation.
We further leverage the explicit distinction in UCCA between types of Scenes (events), applying a specific rule for each of the cases. 
Nevertheless, the DSS approach can be adapted to other semantic schemes,
like AMR \citep{Ba13}. 

We collect human judgments for multiple variants of our system, its sub-components, {\sc Hybrid} and similar systems that use phrase-based MT.
This results in a sizable human evaluation benchmark, which includes 28 systems, totaling at 1960 complex-simple sentence pairs, 
each annotated by three annotators using four criteria.\footnote{The benchmark can be found in \url{https://github.com/eliorsulem/simplification-acl2018}.}
This benchmark will support the future analysis of TS systems, and evaluation practices.

Previous work is discussed in \S\ref{sec:related_work},
the semantic and NMT components we use in  \S\ref{sec:semantics-based} and \S\ref{sec:neural-based} respectively. 
The experimental setup is detailed in \S\ref{sec:experiments}.
Our main results are presented in \S\ref{sec:results}, while \S\ref{sec:additional_exps} presents a more detailed analysis of the system's sub-components and related settings.

\vspace{-0.2cm}
%%%%%%%%%%%%%%%%%%%%%%%%%%%%%%%%%%%%%%%%%%%%%%%%%%%%%%%%%%%%%%%%%%%%%%%%%%%%%%%%%%%
\section{Related Work}\label{sec:related_work}

\paragraph{MT-based sentence simplification.}
Phrase-based Machine Translation \citep[PBMT;][]{K03}
was first used for TS by \citet{S10}, who 
showed good performance on lexical simplification and simple rewriting,
but under-prediction of other operations. 
\citet{Sa15} took a similar approach, finding that it is beneficial to
use training data where the source side is highly similar to the target.
Other PBMT for TS systems include the work of \citet{CK11a}, which uses Moses \citep{K07}, the work of \citet{CK11b}, where the model is extended to include deletion,
and PBMT-R \citep{W12}, where Levenshtein distance to the source is
used for re-ranking to overcome conservatism.

The NTS NMT-based system \citep{Ni17} (henceforth, N17) reported superior performance over PBMT 
in terms of BLEU and human evaluation scores, 
and serves as a component in our system (see Section \ref{sec:neural-based}). 
\citet{Z17} took a similar approach, adding lexical constraints to an NMT model. 
\citet{ZL17} combined NMT with reinforcement learning, using SARI \citep{Xu16}, BLEU, and cosine similarity to the source as the reward.
None of these models explicitly addresses sentence splitting.

\citet{AM17} proposed to reduce conservatism, observed in PBMT and NMT systems, by first identifying simplification 
operations in a parallel corpus and then using sequence-labeling to perform the simplification. However, they did not 
address common structural operations, such as sentence splitting,
and claimed that their method is not applicable to them.

\citet{Xu16} used Syntax-based Machine Translation (SBMT) for sentence simplification,
using a large scale paraphrase dataset \citep{G13} for training. 
While it does not target structural simplification, we include it in our evaluation for completeness.

\vspace{-0.2cm}
\paragraph{Structural sentence simplification.}
Syntactic hand-crafted sentence splitting rules were proposed by \citet{C96}, \citet{S02}, \citet{S11} in the context of rule-based TS. 
The rules separate relative clauses and coordinated clauses and un-embed appositives. In our method, the use of semantic distinctions instead of syntactic ones reduces the number of rules. For example, relative clauses and appositives can correspond 
to the same semantic category. In syntax-based splitting, a generation module is sometimes added after the split \citep{S04}, addressing issues such as
re-ordering and determiner selection. In our model, no explicit regeneration is applied to the split sentences, which are fed directly to an NMT system.

\citet{GS13} used a rule-based system conditioned on event extraction and syntax for defining two simplification models. 
The event-wise simplification one, which separates events to separate output sentences, is similar to our semantic component. 
Differences are in that we use a single semantic representation for defining the rules (rather than a combination
of semantic and syntactic criteria), and avoid the need for complex rules for retaining grammaticality by using
a subsequent neural component.

\vspace{-0.2cm}
\paragraph{Combined structural and lexical TS.}
Earlier TS models used syntactic information for splitting.
\citet{Z10} used syntactic information on the source side, based on the SBMT model of \citet{YK01}.
Syntactic structures were used on both sides in the model of  \citet{WL11}, based on a quasi-synchronous 
grammar \citep{SE06}, which resulted in 438 learned splitting rules.

The model of \citet{SA14} is similar to ours in that it combines linguistic rules for structural simplification and statistical methods for lexical simplification.
However, we use 2 semantic splitting rules instead of their 26 syntactic rules for relative clauses and appositions, and 85 syntactic rules for subordination and coordination.

\citet{NG14} argued that syntactic structures do not always capture the semantic arguments of a frame, 
which may result in wrong splitting boundaries. Consequently, they proposed a supervised system ({\sc Hybrid}) that
uses semantic structures (Discourse Semantic Representations, \citep{K81}) for sentence splitting and deletion.
Splitting candidates are pairs of event variables associated with at least one core thematic 
role (e.g., agent or patient). 
Semantic annotation is used on the source side in both training and test. 
Lexical simplification is performed using the Moses system. 
{\sc Hybrid} is the most similar system to ours architecturally, in that it uses a combination of a semantic structural component and an MT
component. \citet{NG16} proposed instead an unsupervised pipeline, where sentences are split
based on a probabilistic model trained on the semantic structures of Simple Wikipedia as well 
as a language model trained on the same corpus.
Lexical simplification is there performed using the unsupervised model of 
\citet{B11}. As their BLEU and adequacy scores are lower than {\sc Hybrid}'s, 
we use the latter for comparison.

\citet{SG17} combined rule-based simplification conditioned on event extraction, together with an unsupervised lexical simplifier. 
They tackle a different setting, and aim to simplify texts (rather than sentences), by allowing the deletion of entire input sentences.

\vspace{-0.2cm}
\paragraph{Split and Rephrase.}
\citet{N17} recently proposed the Split and Rephrase task, focusing on sentence splitting.
For this purpose they presented a specialized parallel corpus, derived from the WebNLG dataset \citep{G17}.
 The latter is obtained from the DBPedia knowledge base \citep{M12} using content selection and crowdsourcing, and is 
 annotated with semantic triplets of subject-relation-object, obtained semi-automatically.
They experimented with five systems, including one similar to {\sc Hybrid}, 
as well as 
sequence-to-sequence methods for generating sentences from
the source text and its semantic forms.

The present paper tackles both structural and lexical simplification, and examines the effect of sentence splitting on the subsequent application
of a neural system, in terms of its tendency to perform other simplification operations. 
For this purpose, we adopt a semantic corpus-independent approach for sentence splitting that can be easily integrated in 
any simplification system.  
Another difference is that the semantic forms in Split and Rephrase are derived semi-automatically (during corpus compilation), 
while we automatically extract the semantic form, using a UCCA parser.

\begin{figure*}[t!]
  \includegraphics[width=0.48\textwidth]{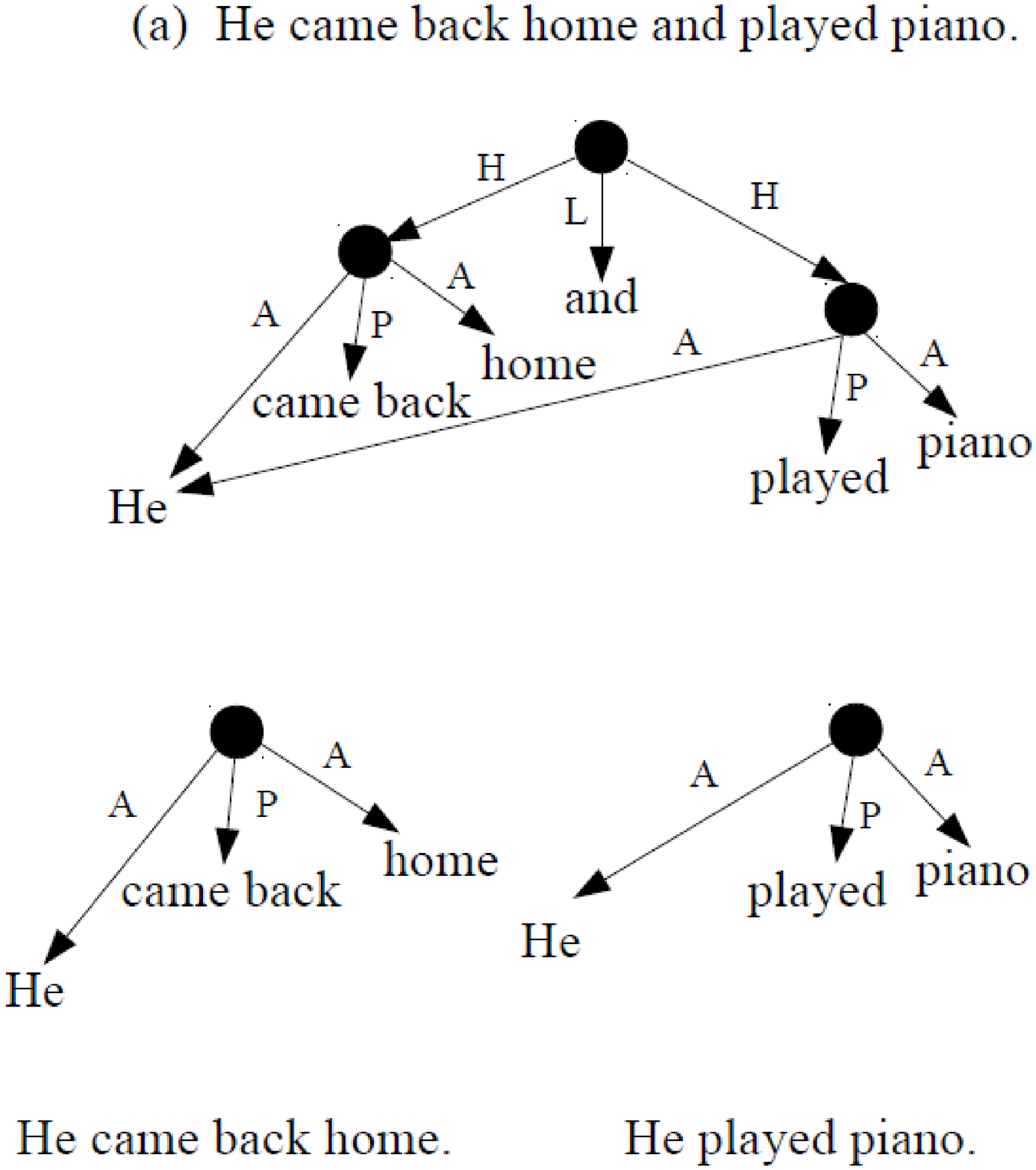}
\hspace{-0.7cm}
\includegraphics[width=0.56\textwidth]{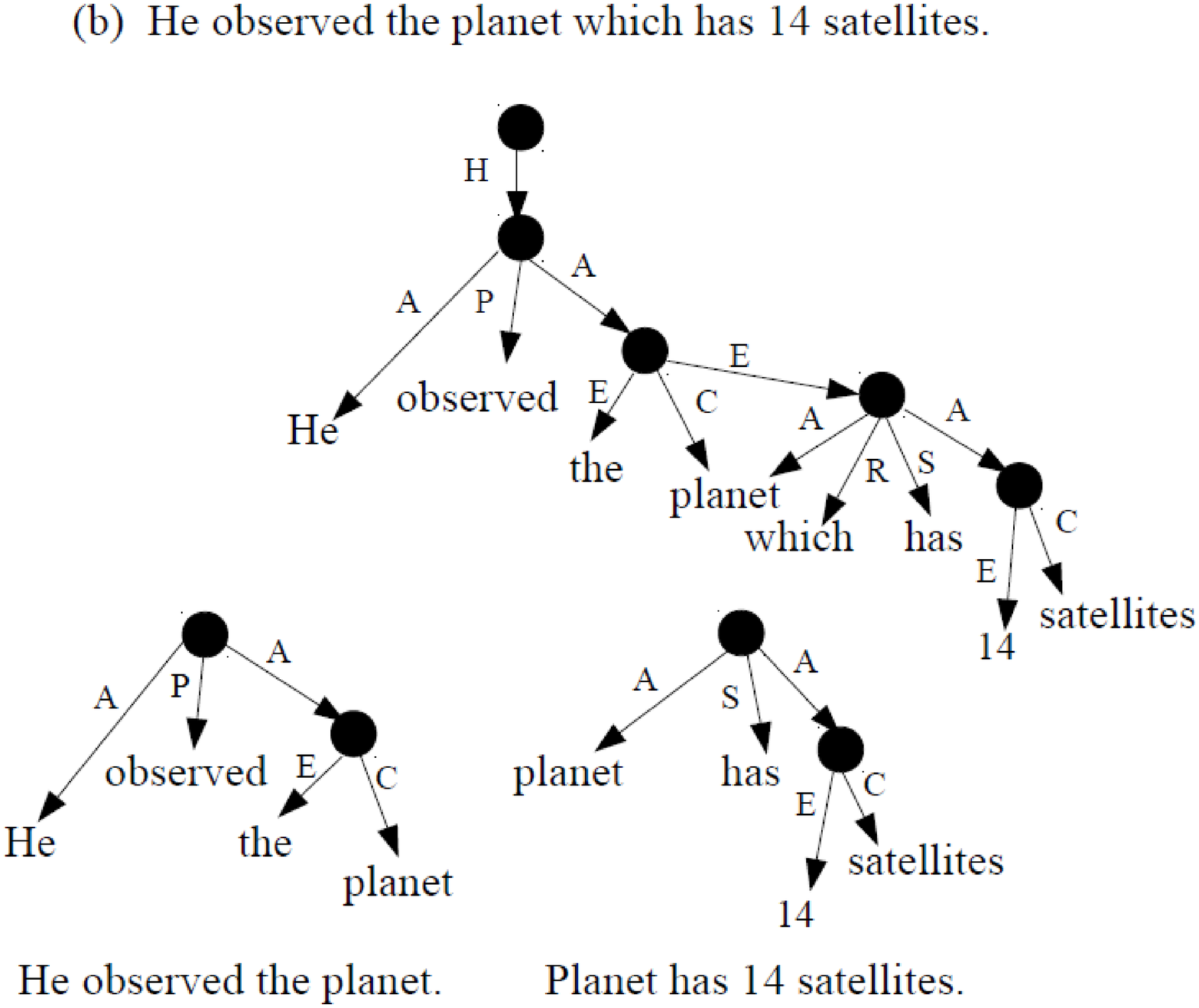}
\caption{Example applications of rules 1 (Figure 1a) and 2 (Figure 1b). In both cases, the original sentence, the semantic parse, the extracted Scenes with the required modifications, and the output of the rules are presented top to bottom.
The UCCA categories used are: Parallel Scene (H), Linker (L), Participant (A), Process/State (P/S), Center (C), Elaborator (E), Relator (R).} 
\label{ucca_rules}
\end{figure*}

\vspace{-0.1cm}
%%%%%%%%%%%%%%%%%%%%%%%%%%%%%%%%%%%%%%%%%%%%%%%%%%%%%%%%%%%%%%%%%%%%%%%%%%%%%%%%%%%%%%%%%%%%%%%%
\section{Direct Semantic Splitting}\label{sec:semantics-based}

\subsection{Semantic Representation}\label{sec:ucca}

UCCA \citep[Universal Cognitive Conceptual Annotation;][]{AR13} is a semantic annotation scheme rooted in typological and cognitive linguistic theory \citep{D10A,D10B,D12,L08}.
It aims to represent the main semantic phenomena in the text, abstracting away from syntactic forms.
UCCA has been shown to be preserved remarkably well across translations \citep{S15} and has also been successfully used
for the evaluation of machine translation \citep{B16} and, recently, for the evaluation of TS \citep{S18} and grammatical error correction \citep{CA18}. 

Formally, UCCA structures are directed acyclic graphs whose nodes (or {\it units}) correspond either to the leaves of the graph %(including the words of the text)%
or to several elements viewed as a single entity according to some semantic or cognitive consideration.

A {\it Scene} is UCCA's notion of an event or a frame, and is a unit that corresponds to a movement, an action or a state which persists in time. 
Every Scene contains one main relation, which can be either a Process or a State. Scenes contain one or more Participants, 
interpreted in a broad sense to include locations and destinations. 
For example, the sentence ``He went to school'' has a single Scene whose Process is ``went''. 
The two Participants  are ``He'' and ``to school''.

Scenes can have several roles in the text. First, they can provide additional information about an established entity 
(Elaborator Scenes), commonly participles or relative clauses. For example, ``(child) who went to school'' is an Elaborator Scene 
in ``The child who went to school is John'' (``child'' serves both as an argument in the Elaborator Scene and as the Center). 
A Scene may also be a Participant in another Scene. For example, ``John went to school'' in the sentence: ``He said John went to school''. 
In other cases, Scenes are annotated as Parallel Scenes (H), which are flat structures and may include a Linker (L), 
as in:  ``When$_L$ [he arrives]$_H$, [he will call them]$_H$''.

With respect to units which are not Scenes, the category Center denotes the semantic head. For example, ``dogs'' is the 
Center of the expression ``big brown dogs'', and ``box'' is the center of ``in the box''. There could be more than one 
Center in a unit, for example in the case of coordination, where all conjuncts are Centers. We define the minimal center of a 
UCCA unit $u$ to be the UCCA graph's leaf reached by starting from $u$ and iteratively selecting the child tagged as Center.

For generating UCCA's structures we use TUPA, a transition-based parser
\citep{H17} (specifically, the TUPA$_{BiLSTM}$ model). TUPA uses an
expressive set of transitions, able to support all structural properties
required by the UCCA scheme. Its transition classifier is based on an MLP
that receives a BiLSTM encoding of elements in the parser state (buffer,
stack and intermediate graph), given word embeddings and other features.

\vspace{-0.1cm}
\subsection{The Semantic Rules} \label{sec:rules}

For performing DSS, we define two simple splitting rules, conditioned on UCCA's categories.
We currently only consider Parallel Scenes and Elaborator Scenes, not 
separating Participant Scenes, in order to avoid splitting in cases of nominalizations or indirect speech.
For example, the sentence ``His arrival surprised everyone'', which has, in addition to the Scene evoked by ``surprised'', a Participant Scene evoked by ``arrival'', is not split here.

\paragraph{Rule \#1.}
Parallel Scenes of a given sentence are extracted, separated in different sentences, 
and concatenated according to the order of appearance. More formally, given a 
decomposition of a sentence $S$ into parallel Scenes $Sc_{1},Sc_{2}, \cdots Sc_{n}$ 
(indexed by the order of the first token), we obtain the following rule, 
where ``$|$'' is the sentence delimiter:
\vspace{-0.6cm}

{\small $$ S \longrightarrow Sc_{1}|Sc_{2}| \cdots |Sc_{n}$$}

\vspace{-0.6cm}
As UCCA allows argument sharing between Scenes, the rule may duplicate the same sub-span of $S$ 
across sentences.
For example, the rule will convert ``He came back home and played piano'' 
into ``He came back home''$|$``He played piano.'' 

%\vspace{-0.2cm}
\paragraph{Rule \#2.}
Given a sentence $S$, the second rule extracts Elaborator Scenes and corresponding minimal centers. 
Elaborator Scenes are then concatenated to the original sentence, where the Elaborator Scenes, 
except for the minimal center they elaborate, are removed. 
Pronouns such as ``who'', ``which'' and ``that'' are also removed. 

Formally, if $\{(Sc_1,C_1) \cdots (Sc_n,C_n)\}$ are the Elaborator Scenes of $S$ 
and their corresponding minimal centers, the rewrite is:
\vspace{-0.3cm}

{\small $$ S \longrightarrow S - \bigcup_{i=1}^{n} (Sc_{i} - C_{i}) | Sc_{1}| \cdots|Sc_{n}$$}
\vspace{-0.4cm}

\noindent
where $S-A$ is $S$ without the unit $A$.
For example, this rule converts the sentence ``He observed the planet which has 14 known satellites'' 
to  ``He observed the planet$|$ Planet has 14 known satellites.''.
Article regeneration is not covered by the rule, as its output is directly fed into the NMT component.

After the extraction of Parallel Scenes and Elaborator Scenes, the resulting simplified Parallel Scenes are placed before the Elaborator Scenes. 
See Figure \ref{ucca_rules}.

\vspace{-0.2cm}
\section{Neural Component}\label{sec:neural-based}
\vspace{-0.1cm}

The split sentences are run through the NTS state-of-the-art neural TS system \citep{Ni17}, 
built using the OpenNMT neural machine translation framework \citep{K17}. 
The architecture includes two LSTM layers, with hidden states of 500 units in each,
as well as global attention combined with input feeding \citep{L15}. Training is done with a 0.3 dropout probability \citep{Sr14}.
This model uses alignment probabilities between the predictions and the original sentences, rather than character-based models, to retrieve the original words.

We here consider the w2v initialization for NTS (N17), where word2vec embeddings of size 300 are trained on Google News \citep{M13a} and local embeddings of size 200 are trained on the training simplification corpus \citep{RS10, M13b}.
Local embeddings for the encoder are trained on the source side of the training corpus, while those for the decoder are trained on the simplified side.

For sampling multiple outputs from the system, beam search is performed during decoding by generating the first 5 hypotheses at each step ordered by the 
log-likelihood of the target sentence given the input sentence.
We here explore both the highest (h1) and fourth-ranked (h4) hypotheses, which we show to increase the SARI score and to be much less conservative.\footnote{Similarly, N17
considered the first two hypotheses and showed that h2 has an higher SARI score and is less conservative than h1.}
We thus experiment with two variants of the neural component, denoted by NTS-h1   and NTS-h4.
The pipeline application of the rules and the neural system results in two corresponding models: SENTS-h1 and SENTS-h4.

\vspace{-0.2cm}
\section{Experimental Setup}\label{sec:experiments}

\paragraph{Corpus}
All systems are tested on the test corpus of \citet{Xu16},\footnote{\url{https://github.com/cocoxu/simplification} (This also includes SARI tools and the SBMT-SARI system.)} comprising 359 sentences from the PWKP corpus \citep{Z10} with 8 references collected by crowdsourcing for each of the sentences. 
\vspace{-0.3cm}

\paragraph{Semantic component.}
The TUPA parser\footnote{\url{ https://github.com/danielhers/tupa}} is trained on the UCCA-annotated {\it Wiki} corpus.\footnote{\url{ http://www.cs.huji.ac.il/~oabend/ucca.html}}
\vspace{-0.2cm}
\paragraph{Neural component.}
We use the NTS-w2v model\footnote{\url{https://github.com/senisioi/NeuralTextSimplification}} provided by N17, obtained by training on the corpus of \citet{H15} and tuning on the corpus of \citet{Xu16}. The training set is based on manual and automatic alignments between standard English Wikipedia and Simple English Wikipedia, including both good matches and partial matches whose similarity score is above the 0.45 scale threshold \citep{H15}. The total size of the training set is about 280K aligned sentences, of
which 150K sentences are full matches and 130K are partial matches.\footnote{We also considered the default initialization for the neural component, using the NTS model without word embeddings. Experimenting on the tuning set, the w2v approach got higher BLEU and SARI scores (for h1 and h4 respectively) than the default approach.}

\vspace{-0.2cm}
\paragraph{Comparison systems.}
We compare our findings to {\sc Hybrid}, which is the state of the art for joint structural and lexical simplification, 
implemented by \citet{ZL17}.\footnote{\url{ https://github.com/XingxingZhang/dress}}
We use the released output of {\sc Hybrid}, trained on a corpus extracted from Wikipedia, which includes the aligned sentence pairs from \citet{K13}, 
the aligned revision sentence pairs in \citet{WL11}, and the PWKP corpus, totaling about 296K sentence pairs. 
The tuning set is the same as for the above systems.

In order to isolate the effect of NMT, we also implement SEMoses, where the neural-based component is replaced by the phrase-based MT 
system Moses,\footnote{\url{ http://www.statmt.org/moses/}} which is also used in {\sc Hybrid}. 
The training, tuning and test sets are the same as in the case of SENTS. 
MGIZA\footnote{\url{ https://github.com/moses-smt/mgiza}} is used for word alignment. 
The KenLM language model is trained using the target side of the training corpus.

\vspace{-.2cm}
\paragraph{Additional baselines.}
We report human and automatic evaluation scores for Identity (where the output is identical to the input), for Simple Wikipedia where the output is the corresponding aligned sentence in the PWKP corpus, and for the SBMT-SARI system, 
tuned against SARI \citep{Xu16}, which 
maximized the SARI score on this test set in previous works \citep{Ni17,ZL17}.

\vspace{-0.2cm}
\paragraph{Automatic evaluation.}\label{sec:automatic_evaluation}
The automatic metrics used for the evaluation are: 
(1) BLEU \citep{P02}
(2) SARI \citep[System output Against References and against the Input sentence;][]{Xu16}, which compares the n-grams of the system output with those
of the input and the human references, separately evaluating  the  quality  of  words  that  are  added, deleted and kept by the systems. 
(3) $F_{\rm add}$: the addition component of the SARI score (F-score);
(4) $F_{\rm keep}$: the keeping component of the SARI score (F-score); (5) {\bf $P_{\rm del}$}: the deletion component of the SARI score (precision).\footnote{Uniform tokenization and truecasing styles for all systems are obtained using the Moses toolkit.}
Each metric is computed against the 8 available references.
We also assess system conservatism, reporting the percentage of sentences copied from the input (\%Same), the averaged Levenshtein distance from the source (${\rm LD_{SC}}$, which considers additions, deletions, and substitutions), and the number of source sentences that are split ($\#$Split).\footnote{We used the NLTK package \citep{LB02} for these computations.}

\vspace{-0.2cm}
\paragraph{Human evaluation.} \label{sec:human_evaluation}
Human evaluation is carried out by 3 in-house native English annotators, 
who rated the different input-output pairs for the different systems according to 4 parameters: Grammaticality (G), 
Meaning preservation (M), Simplicity (S) and Structural Simplicity (StS). Each input-output pair is rated by all 3 annotators.
Elicitation questions are given in Table \ref{tab:hum_eval}.

As the selection process of the input-output pairs in the test corpus of \citet{Xu16}, as well as their crowdsourced references, are explicitly biased towards lexical simplification, the use of human evaluation permits us to evaluate the structural aspects of the system outputs, even where structural operations are not attested in the references. Indeed, we show that system outputs may receive considerably higher structural simplicity scores than the source, in spite of the sample selection bias.

Following previous work \citep[e.g.,][]{NG14,Xu16,Ni17}, Grammaticality (G) and Meaning preservation (M) are measured using a 1 to 5 scale. 
Note that in the first question, the input sentence is not taken into account. The grammaticality of the input is assessed by 
evaluating the Identity transformation (see Table \ref{human_evaluation}), providing a baseline for the grammaticality scores of the other systems.

Following N17, a -2 to +2 scale is used for measuring simplicity, where a 0 score indicates that the input and the output are equally complex. This scale, compared to the standard 1 to 5 scale, permits a better differentiation between cases where simplicity is hurt (the output is more complex than the original) and between cases where the output is as simple as the original, for example in the case of the identity transformation.
Structural simplicity is also evaluated with a -2 to +2 scale. 
The question for eliciting StS is accompanied with a negative example, showing a case of lexical simplification, where a complex word 
is replaced by a simple one (the other questions appear without examples). 
A positive example is not included so as not to bias the annotators by revealing the nature of the operations we focus on
(splitting and deletion). 
We follow N17 in applying human evaluation on the first 70 sentences of the test corpus.\footnote{We do not exclude system outputs identical to the source, as done by N17.}

The resulting corpus, totaling 1960 sentence pairs, each annotated by 3 annotators, also include the additional experiments described in Section \ref{sec:additional_exps} as well as the outputs of the NTS and SENTS systems used with the default initialization.

The inter-annotator agreement, using Cohen's quadratic weighted $\kappa$ \citep{C68}, is computed as the average agreement of the 3 annotator pairs.
The obtained rates are 0.56, 0.75, 0.47 and 0.48 for G, M, S and StS respectively.

System scores are computed by averaging over the 3 annotators and the 70 sentences.

\vspace{-0.1cm}
\begin{center}
\begin{table}[h]
\small
\begin{center}
\begin{tabular}{|l|l|}
\hline
G & \parbox{0.75\linewidth}{Is the output fluent and grammatical?}\\
\hline
M & \parbox{0.75\linewidth}{Does the output preserve the meaning of the input?}\\
\hline
S &\parbox{0.75\linewidth}{Is the output simpler than the input?}\\
\hline
StS &\parbox{0.75\linewidth}{Is the output simpler than the input, ignoring the complexity of the words?}\\
\hline
\end{tabular}
\end{center}
\hfill
\vspace{-0.2cm}
\caption{\label{tab:hum_eval} Questions for the human evaluation.}
\label{questions} 
\end{table}
\vspace{-0.5cm}
\end{center}

\begin{center}
\begin{table}[ht]
\scriptsize
\centering
\begin{tabular}{|c|c|c|c|c|}
\hline
& {\bf G} & {\bf M} & {\bf S} & {\bf StS} \\
\hline\hline
{\bf Identity} & 4.80 &  5.00 & 0.00 & 0.00 \\
\hline    
{\bf Simple Wikipedia} & 4.60 & 4.21 &  0.83 &  0.38 \\
\hline\hline
\multicolumn{5}{|c|}{{\bf Only MT-Based Simplification}}\\
\hline
{\bf SBMT-SARI} & 3.71& 3.96& 0.14& -0.15 \\
\hline
{\bf NTS-h1} & 4.56 & {\bf 4.48} & 0.22 & 0.15 \\

\hline
{\bf NTS-h4} & 4.29 & 3.90 & 0.31 & 0.19 \\
\hline\hline
\multicolumn{5}{|c|}{{\bf Only Structural Simplification}}\\
\hline
{\bf DSS}& 3.42& 4.15&  0.16& 0.16 \\
\hline\hline
\multicolumn{5}{|c|}{{\bf Structural+MT-based Simplification}}\\
\hline
{\bf Hybrid} & 2.96 & 2.46 & 0.43 & 0.43 \\
\hline
{\bf SEMoses}& 3.27 & 3.98 & 0.16 & 0.13 \\
\hline
{\bf SENTS-h1} & 3.98 & 3.33 & {\bf 0.68} & {\bf 0.63} \\
\hline
{\bf SENTS-h4} & 3.54 & 2.98 & 0.50 & 0.36 \\
\hline
\end{tabular}
\hfill
\caption{Human evaluation of the different NMT-based systems. Grammaticality (G) and Meaning preservation (M) are measured using a 1 to 5 scale. A -2 to +2 scale is used for measuring simplicity (S) and structural simplicity (StS) of the output relative to the input sentence. The highest score in each column appears in bold.
Structural simplification systems are those that explicitly model structural operations.}
\label{human_evaluation}
\end{table}
\end{center}
\vspace{-1.02cm}

\begin{center}
\begin{table*}[t]
\scriptsize
\centering
\begin{tabular}{|c|c|c|c|c|c||c|c|c|}
\hline
 & {\bf BLEU} & {\bf SARI} & {\bf $F_{\rm add}$} & {\bf $F_{\rm keep}$} & {\bf $P_{\rm del}$} & {\bf $\%$ Same} & ${\rm LD_{SC}}$ & $\#$Split\\
\hline\hline
{\bf Identity} &  94.93 &25.44 & 0.00 &  76.31 & 0.00 & 100 & 0.00 & 0\\
\hline
{\bf Simple Wikipedia} & 69.58  &  39.50  &  8.46 & 61.71 &  48.32 & 0.00 & 33.34 & 0\\
 \hline\hline
 \multicolumn{9}{|c|}{{\bf Only MT-Based Simplification}}\\
\hline
{\bf SBMT-SARI} & 74.44 & {\bf 41.46} & {\bf 6.77} & 69.92 & {\bf 47.68} & 4.18 & 23.31 & 0\\
\hline
{\bf NTS-h1} &  {\bf 88.67} & 28.73 & 0.80 & {\bf 70.95} & 14.45 & 66.02 & 17.13 & 0\\
\hline
{\bf NTS-h4}&79.88 & 36.55 & 2.59 & 65.93 &  41.13 &2.79 & 24.18 & 1\\
\hline\hline
\multicolumn{9}{|c|}{{\bf Only Structural Simplification}}\\
\hline 
{\bf DSS} &76.57 &  36.76 & 3.82 & 68.45 & 38.01 & 8.64 & 25.03 & 208\\
\hline\hline
\multicolumn{9}{|c|}{{\bf Structural+MT-Based Simplification}}\\
\hline
{\bf\sc Hybrid} & 52.82 & 27.40 & 2.41 & 43.09 & 36.69 & 1.39 & 61.53 & 3\\
\hline
{\bf SEMoses} &74.45  &36.68 & 3.77 & 67.66 & 38.62 & 7.52 & 27.44 & 208\\
\hline    
{\bf SENTS-h1} &  58.94 & 30.27 & 3.01 & 51.52 & 36.28 & 6.69 & 59.18 & 0\\
\hline
{\bf SENTS-h4}& 57.71 &  31.90 & 3.95 &  51.86 & 39.90 & 0.28 & 54.47 & 17\\
\hline   
\end{tabular}
\vspace{-0.1cm}
\hfill
\caption{The left-hand side of the table presents BLEU and SARI scores for the combinations of NTS and DSS, as well as for the baselines.
The highest score in each column appears in bold. The right hand side presents lexical and structural properties of the outputs.
$\%$Same: proportion of sentences copied from the input; ${\rm LD_{SC}}$: Averaged Levenshtein distance from the source; $\#$Split: number of split sentences.
Structural simplification systems are those that explicitly model structural operations.}
\label{automatic_evaluation}
\end{table*}
\end{center}

\begin{center}
\begin{table*}[ht]
\scriptsize
\centering
\begin{tabular}{|c|c|c|c|c|c||c|c|c||c|c|c|c|}
\hline
& {\bf BLEU} & {\bf SARI} & {\bf $F_{\rm add}$} & {\bf $F_{\rm keep}$} & {\bf $P_{\rm del}$} & {\bf $\%$ Same} & ${\rm LD_{SC}}$ & $\#$Split &{\bf G} & {\bf M} & {\bf S} & {\bf StS} \\
\hline
{\bf Moses} &92.58 &28.19 & 0.16 & 75.73 & 8.70 & 79.67 & 3.22& 0&4.25 & 4.78 & 0 & 0.04 \\
\hline      
{\bf SEMoses} &74.45  &36.68 & 3.77 & 67.66 & 38.62 & 7.52 & 27.44&208 &3.27 & 3.98 & {\bf 0.16} & 0.13 \\
\hline
{\bf SETrain1-Moses} &91.24 &33.06 & 0.41 & 76.07 & 22.69 &60.72 &4.47& 1& 4.23& 4.54& -0.12& -0.13 \\
\hline     
{\bf SETrain2-Moses} & {\bf 94.31} &26.71 & 0.07 &  {\bf 76.20}&3.85 &92.76&1.45& 0&4.73 & {\bf 4.99} & 0.01 & -0.005 \\
\hline
{\bf Moses$_{{\rm LM}}$} &92.66 &28.19 & 0.18 & 75.68 & 8.71 & 79.39 & 3.43& 0 & 4.55 &  4.82 & -0.01 & -0.04 \\
\hline
{\bf SEMoses$_{{\rm LM}}$} &74.49 &  {\bf 36.70} &  {\bf 3.79} & 67.67 &  {\bf 38.65}& 7.52 &27.45& 208 & 3.32 & 4.08  & 0.15 &  {\bf 0.14}\\
\hline
{\bf SETrain1-Moses$_{{\rm LM}}$} &85.68 &36.52 & 2.34 & 72.85 & 34.37 & 27.30 &6.71& 33& 4.03 & 4.63 & -0.11 & -0.12 \\
\hline
{\bf SETrain2-Moses$_{{\rm LM}}$} &94.22 &26.66 & 0.10 & 76.19 & 3.69& 92.20 &1.43& 0&  {\bf 4.75} & {\bf 4.99} & 0.01 & -0.01\\
\hline
\end{tabular}
\vspace{-0.1cm}
\hfill
\caption{Automatic and human evaluation for the different combinations of Moses and DSS. %The highest score in each column appears in bold.
The automatic metrics as well as the lexical and structural properties reported ($\%$Same: proportion of sentences copied from the input; ${\rm LD_{SC}}$: Averaged Levenshtein distance from the source; $\#$Split: number of split sentences) concern the 359 sentences of the test corpus. Human evaluation, with the G, M, S, and StS parameters, is applied to the first 70 sentences of the corpus.
The highest score in each column appears in bold.
}

\label{moses-based}
\end{table*}
\end{center}

%%%%%%%%%%%%%%%%%%%%%%%%%%%%%%%%%%%%%%%%%%%%%%%%%%%%%%%%%%%%%%%%%%%%%%%%%%%%%%%%%%%%%%%%%%%%%%%%%%%%%%%%%%%%%%%%%%%%%%%%%%%%%%%%%%%%%%%%%%%%%%%%%%%%%%%%%%%%%%%%%%%%%%%%%%%%%%
\vspace{-1.6cm}
\section{Results} \label{sec:results}

\paragraph{Human evaluation.}
Results are presented in Table \ref{human_evaluation}. 
First, we can see that the two SENTS systems outperform {\sc Hybrid} in terms of G, 
M, and S. SENTS-h1 is the best scoring system, under all human measures.

In comparison to NTS, SENTS scores markedly higher on the simplicity judgments.
Meaning preservation and grammaticality are lower for SENTS, which is likely due to the more
conservative nature of NTS. Interestingly, the application of the splitting rules by themselves
does not yield a considerably simpler sentence. 
This likely stems from the rules not necessarily yielding grammatical sentences
(NTS often serves as a grammatical error corrector over it), and from the incorporation
of deletions, which are also structural operations, and are performed by the neural system.

An example of high structural simplicity scores for SENTS resulting from deletions is presented in Table \ref{tab:output_example}, together with the outputs of the other systems and the corresponding human evaluation scores.
NTS here performs  lexical simplification, replacing the word ``incursions'' by ``raids'' or ``attacks'''.
On the other hand, the high StS scores  obtained by DSS and SEMoses are due to sentence splittings.

\vspace{-0.2cm}
\paragraph{Automatic evaluation.}
Results are presented in Table \ref{automatic_evaluation}.
Identity obtains much higher BLEU scores than any other system, suggesting that BLEU may not be informative in this setting. 
SARI seems more informative, and assigns the lowest score to Identity and the second highest to the reference.

Both SENTS systems outperform {\sc Hybrid} in terms of SARI and all its 3 sub-components. 
The h4 setting (hypothesis \#4 in the beam) is generally best,
both with and without the splitting rules.

Comparing SENTS to using NTS alone (without splitting), 
we see that SENTS obtains higher SARI scores when hypothesis \#1 is used and that NTS obtains 
higher scores when hypothesis \#4 is used.
This may result from NTS being more conservative than SENTS (and {\sc Hybrid}), which is rewarded by SARI
(conservatism is indicated by the $\%$Same column). Indeed for h1, \%Same is reduced 
from around 66$\%$ for NTS, to around 7$\%$ for SENTS. 
Conservatism further decreases when h4 is used (for both NTS and SENTS).
Examining SARI's components, we find that SENTS outperforms NTS on $F_{add}$,
and is comparable (or even superior for $h1$ setting) to NTS on $P_{del}$. The superior SARI
score of NTS over SENTS is thus entirely a result of a superior $F_{keep}$, which is easier for a conservative system to maximize. 

Comparing {\sc Hybrid} with SEMoses, both of which use Moses, 
we find that SEMoses obtains higher BLEU and SARI scores, as well as G and M human scores, and splits many more sentences.
{\sc Hybrid} scores higher on the human simplicity measures. We note, however, that applying NTS alone is inferior to {\sc Hybrid}
in terms of simplicity, and that both components are required to obtain high simplicity scores (with SENTS).

We also compare the sentence splitting component used in our systems (namely DSS) to that used in {\sc Hybrid}, abstracting away from deletion-based and lexical simplification. We therefore apply DSS to the test set (554 sentences) of the WEB-SPLIT corpus \citep{N17} (See Section \ref{sec:related_work}), which focuses on sentence splitting. We compare our results to those reported for a variant of {\sc Hybrid} used without the deletion module, and trained on WEB-SPLIT \citep{N17}. DSS gets a higher BLEU score (46.45 vs. 39.97) and performs more splittings (number of output sentences per input sentence of 1.73 vs. 1.26).

\vspace{-0.5cm}
\begin{center}
\begin{table*}[h]
\scriptsize
\begin{center}
\setlength\tabcolsep{1pt} 
\begin{tabular}{|c|l|c|c|c|c|}
\hline
& & {\bf G} & {\bf M} & {\bf S} & {\bf StS} \\
\hline
{\bf Identity}& \parbox{0.75\linewidth}{\vspace{0.05cm} In​ return,​ Rollo​ swore​ fealty​ to​ ​Charles,​ ​converted​ ​to​ Christianity,​ ​and
​undertook​ ​to​ ​defend​ ​the​ ​northern​ ​region​ ​of​ ​France​ ​against​ ​the​ ​incursions​ ​of​ ​other
​Viking​ ​groups. \vspace{0.1cm}} & 5.00& 5.00& 0.00& 0.00\\
\hline    
{\bf Simple Wikipedia}& \parbox{0.75\linewidth}{\vspace{0.05cm} In​ return,​ Rollo​ swore​ fealty​ to​ Charles,​ ​converted​ to​ Christianity,​ ​and​ swore​ ​to
  ​defend​ ​the​ northern​ region​ of​ France​ ​against​ raids​ by​ other​ ​Viking​ groups. \vspace{0.1cm}} & 4.67 & 5.00& 1.00& 0.00\\
\hline
{\bf SBMT-SARI}& \parbox{0.75\linewidth}{\vspace{0.02cm} In​ ​return,​ ​Rollo​ ​swore​ ​fealty​ ​to​ ​Charles,​ converted​ to​ ​Christianity,​ and​ set​ out​ to
  ​defend​ ​the​ north​ of​ France​ from​ ​the​ ​raids​ ​of​ ​other​ ​viking​ ​groups. \vspace{0.1cm}} & 4.67 & 4.67 & 0.67 & 0.00\\
\hline
{\bf NTS-h1}& \parbox{0.75\linewidth}{In​ ​return,​ ​Rollo​ ​swore​ ​fealty​ ​to​ ​Charles,​ ​converted​ ​to​ ​Christianity,​ and​ ​undertook
​to​ ​defend​ the​ ​northern​ ​region​ of​ France​ ​against​ ​the​ raids​ ​of​ ​other​ ​Viking​ groups. \vspace{0.1cm}} & 5.00& 5.00& 1.00& 0.00\\
\hline
{\bf NTS-h4}& \parbox{0.75\linewidth}{​In​ ​return,​ Rollo​ ​swore​ fealty​ ​to​ ​Charles,​ ​converted​ ​to​ ​Christianity,​ ​and​ ​undertook
​to​ ​defend​ ​the​ ​northern​ ​region​ ​of​ ​France​ ​against​ ​the​ ​attacks​ ​of​ ​other​ ​Viking​ ​groups. \vspace{0.1cm}} & 4.67 & 5.00& 1.00& 0.00\\
\hline
{\bf DSS} & \parbox{0.75\linewidth}{​Rollo​ ​swore​ ​fealty​ ​to​ ​Charles.​ ​Rollo​ ​converted​ ​to​ ​Christianity.​ ​Rollo​ ​undertook​ ​to
​defend​ ​the​ ​northern​ ​region​ ​of​ ​France​ ​against​ the​ ​incursions​ ​of​ ​other​ ​viking​ ​groups. \vspace{0.1cm}} & 4.00& 4.33 & 1.33 & 1.33\\
\hline
{\bf\sc Hybrid}& \parbox{0.75\linewidth}{In​ ​return​ ​Rollo​ ​swore,​ ​and​ ​undertook​ ​to​ ​defend​ ​the​ ​region​ ​of​ ​France.,​ ​Charles,
​converted \vspace{0.1cm}} & 2.33 & 2.00& 0.33 & 0.33\\
\hline
{\bf SEMoses} & \parbox{0.75\linewidth}{\vspace{0.02cm} Rollo swore put his seal to Charles. Rollo converted to Christianity. Rollo
undertook to defend the northern region of France against the incursions of other viking
groups. \vspace{0.1cm}} & 3.33 & 4.00 & 1.33 & 1.33\\
\hline
{\bf SENTS-h1} & \parbox{0.75\linewidth}{\vspace{0.02cm} Rollo​ swore​ ​fealty​ ​to​ ​Charles. \vspace{0.1cm}} & 5.00 & 2.00 & 2.00 & 2.00\\
\hline
{\bf SENTS-h4} & \parbox{0.75\linewidth}{\vspace{0.02cm} Rollo​ swore​ ​fealty​ ​to​ ​Charles​ ​and​ ​converted​ ​to​ ​Christianity.} & 5.00& 2.67 & 1.33 & 1.33\\
\hline
\end{tabular}
\end{center}
\vspace{-0.2cm}
\hfill
\caption{\label{tab:output_example} System outputs for one of the test sentences with the corresponding human evaluation scores (averaged over the 3 annotators). Grammaticality (G) and Meaning preservation (M) are measured using a 1 to 5 scale. A -2 to +2 scale is used for measuring simplicity (S) and structural simplicity (StS) of the output relative to the input sentence.}
\label{questions} 
\end{table*}
\vspace{-0.2cm}
\end{center}

\vspace{-0.3cm}
\begin{center}
\begin{table}[t]
\scriptsize
\centering
\begin{tabular}{|c|c|c|c|c|}
\hline
& {\bf G} & {\bf M} & {\bf S} & {\bf StS} \\
\hline\hline
{\bf DSS$^m$}& 3.38& 3.91& -0.16& -0.16 \\
\hline
{\bf SENTS$^m$-h1} & 4.12 & 3.34 & 0.61 & 0.58 \\
\hline
{\bf SENTS$^{m}$-h4} & 3.60 & 3.24 & 0.26 & 0.12 \\
\hline
{\bf SEMoses}$^{m}$ & 3.32& 4.27& -0.25& -0.25 \\
\hline
{\bf SEMoses}$_{LM}^{m}$ & 3.43& 4.28 & -0.18& -0.19 \\
\hline    
\end{tabular}
\vspace{-0.1cm}
\hfill
\caption{\label{tab:human_evaluation_semisuper}
Human evaluation using manual UCCA annotation. Grammaticality (G) and Meaning preservation (M) are measured using a 1 to 5 scale. A -2 to +2 scale is used for measuring simplicity (S) and structural simplicity (StS) of the output relative to the input sentence. X$^m$ refers to the semi-automatic version of the system X.
}
\end{table}
\end{center}

\section{Additional Experiments} \label{sec:additional_exps}

\paragraph{Replacing the parser by manual annotation.}
In order to isolate the influence of the parser on the results, we implement a semi-automatic version of the semantic component, 
which uses manual UCCA annotation instead of the parser, focusing of the first 70 sentences of the test corpus. 
We employ a single expert UCCA annotator and use the UCCAApp annotation tool \cite{A17}.

Results are presented in Table \ref{tab:human_evaluation_semisuper}, for both SENTS and SEMoses. 
In the case of SEMoses, meaning preservation is improved when manual UCCA annotation is used. 
On the other hand, simplicity degrades, possibly due to the larger number of Scenes marked by the human annotator
(TUPA tends to under-predict Scenes).
This effect doesn't show with SENTS, where trends are similar to the automatic parses case, and high simplicity scores are obtained.
This demonstrates that UCCA parsing technology is sufficiently mature to be used to carry out structural simplification.

We also directly evaluate the performance of the parser by computing F1, Recall and Precision DAG scores \citep{H17}, against the manual UCCA annotation.\footnote{We use the evaluation tools provided in \url{https://github.com/danielhers/ucca}, ignoring 9 sentences for which different tokenizations of proper nouns are used in the automatic and manual parsing.} We obtain for primary edges (i.e. edges that form a tree structure) scores of 68.9 $\%$, 70.5$\%$, and 67.4$\%$ for F1, Recall and Precision respectively. For remotes edges (i.e. additional edges, forming a DAG), the scores are 45.3$\%$, 40.5$\%$, and 51.5$\%$. These results are comparable with the  out-of-domain results reported by \citet{H17}.

\vspace{-0.2cm}
\paragraph{Experiments on Moses.}

We test other variants of SEMoses, where phrase-based MT is used instead of NMT.
Specifically, we incorporate semantic information in a different manner by implementing two additional models: 
(1) SETrain1-Moses, where a new training corpus is obtained by applying the splitting rules to the target 
side of the training corpus; (2) SETrain2-Moses, where the rules are applied to the source side. 
The resulting parallel corpus is concatenated to the original training corpus.
We also examine whether training a language model (LM) on split sentences has a positive effect,
and train the LM on the split target side.
For each system $X$, the version with the LM trained on split sentences is denoted by $X_{LM}$.

We repeat the same human and automatic evaluation protocol as in \S\ref{sec:results},
presenting results in Table \ref{moses-based}. 
Simplicity scores are much higher in the case of SENTS (that uses NMT), than with Moses.
The two best systems according to SARI are SEMoses and SEMoses$_{LM}$ which use DSS. 
In fact, they resemble the performance of DSS applied alone (Tables \ref{human_evaluation} and \ref{automatic_evaluation}),
which confirms the high degree of conservatism observed by Moses in simplification \citep{AM17}.
Indeed, all Moses-based systems that don't apply DSS as
pre-processing are conservative, obtaining high scores for BLEU, grammaticality and meaning preservation,
but low scores for simplicity.
Training the LM on split sentences shows little improvement.

\vspace{-0.2cm}
\section{Conclusion} \label{sec:conclusion}
\vspace{-0.1cm}
We presented the first simplification system combining semantic structures and neural machine translation, showing that it outperforms
existing lexical and structural systems.
The proposed approach addresses the over-conservatism of MT-based systems for TS,
which often fail to modify the source in any way.
The semantic component performs sentence splitting without relying on a specialized corpus, 
but only an off-the-shelf semantic parser. The consideration of sentence splitting as a decomposition of a sentence into its Scenes is further supported by recent work on structural TS evaluation \citep{S18}, which proposes the \textsf{SAMSA} metric. The two works, which apply this assumption to different ends (TS system construction, and TS evaluation), confirm its validity.
Future work will leverage UCCA's cross-linguistic applicability to support multi-lingual TS and TS pre-processing for MT.

\vspace{-0.1cm}
\section*{Acknowledgments}
\vspace{-0.1cm}
We would like to thank Shashi Narayan for sharing his data and the annotators for participating in our evaluation and UCCA annotation experiments.
We also thank Daniel Hershcovich and the anonymous reviewers for their helpful advices. This work was partially supported by the Intel Collaborative Research Institute for Computational Intelligence (ICRI-CI) and by the Israel Science Foundation
(grant No. 929/17), as well as by the HUJI Cyber Security Research
Center in conjunction with the Israel National Cyber
Bureau in the Prime Minister's Office.

%\bibliography{acl2018}
\bibliography{bibliosimplification}

\begin{thebibliography}{64}
\expandafter\ifx\csname natexlab\endcsname\relax\def\natexlab#1{#1}\fi

\bibitem[{Abend and Rappoport(2013)}]{AR13}
Omri Abend and Ari Rappoport. 2013.
\newblock {{U}niversal
  {C}onceptual {C}ognitive {A}nnotation ({UCCA})}.
\newblock In \emph{Proc. of ACL-13}, pages 228--238.
\newblock \url{http://aclweb.org/anthology/P/P13/P13-1023.pdf}.

\bibitem[{Abend et~al.(2017)Abend, Yerushalmi, and Rappoport}]{A17}
Omri Abend, Shai Yerushalmi, and Ari Rappoport. 2017.
\newblock {{UCCAA}pp:
  {W}eb-application for syntactic and semantic phrase-based annotation}.
\newblock In \emph{Proc. of ACL'17, System Demonstrations}, pages 109--114.
\newblock \url{http://www.aclweb.org/anthology/P17-4019}.

\bibitem[{Alu\'{i}sio and Gasperin(2010)}]{AG10}
Sandra~Maria Alu\'{i}sio and Caroline Gasperin. 2010.
\newblock {Foestering
  disgital inclusion and accessibility: {T}he {P}or{S}imples project for
  simplification of {P}ortuguese texts}.
\newblock In \emph{Proc. of NAACL HLT 2010 Young Investigators Workshop on
  Computational Approaches to Languages of the Americas}, pages 46--53.
\newblock \url{http://anthology.aclweb.org/W/W10/W10-1607.pdf}.

\bibitem[{Alva-Manchego et~al.(2017)Alva-Manchego, Bingel, Paetzold, Scarton,
  and Specia}]{AM17}
Fernando Alva-Manchego, Joachim Bingel, Gustavo~H. Paetzold, Carolina Scarton,
  and Lucia Specia. 2017.
\newblock {Learning how to
  simplify from explicit labeling of complex-simplified text pairs}.
\newblock In \emph{Proc. of IJCNLP'17}, pages 295--305.
\newblock \url{http://aclweb.org/anthology/I17-1030}.

\bibitem[{Banarescu et~al.(2013)Banarescu, Bonial, Cai, Georgescu, Griffitt,
  Hermjakob, Knight, Koehn, Palmer, and Schneider}]{Ba13}
Laura Banarescu, Claire Bonial, Shu Cai, Madalina Georgescu, Kira Griffitt, Ulf
  Hermjakob, Kevin Knight, Philipp Koehn, Martha Palmer, and Nathan Schneider.
  2013.
\newblock {{A}bstract
  {M}eaning {R}representation for sembanking}.
\newblock \emph{Proc. of Linguistic Annotation Workshop and Interoperability
  with Discourse}, pages 178--186.
\newblock \url{http://aclweb.org/anthology/W/W13/W13-2322.pdf}.

\bibitem[{Barlacchi and Tonelli(2013)}]{BT13}
Gianni Barlacchi and Sara Tonelli. 2013.
\newblock {ERNESTA}: {A} sentence simplification tool for children's stories in
  {I}talian.
\newblock In \emph{Proc. of CICLing'13}, pages 476--487.

\bibitem[{Biran et~al.(2011)Biran, Brody, and Elhadad}]{B11}
Or~Biran, Samuel Brody, and No\'{e}mie Elhadad. 2011.
\newblock {Putting it
  simply: a context-aware approach to lexical simplification}.
\newblock In \emph{Proc. of ACL'11}, pages 465--501.
\newblock \url{http://aclweb.org/anthology/P/P11/P11-2087.pdf}.

\bibitem[{Birch et~al.(2016)Birch, Abend, Bojar, and Haddow}]{B16}
Alexandra Birch, Omri Abend, Ond\v{r}ej Bojar, and Barry Haddow. 2016.
\newblock {{HUME}:
  {H}uman {UCCA}-based evaluation of machine translation}.
\newblock In \emph{Proc. of EMNLP'16}, pages 1264--1274.
\newblock \url{http://aclweb.org/anthology/D/D16/D16-1134.pdf}.

\bibitem[{Chandrasekar et~al.(1996)Chandrasekar, Doran, and Srinivas}]{C96}
Raman Chandrasekar, Christine Doran, and Bangalore Srinivas. 1996.
\newblock {Motivations
  and methods for sentence simplification}.
\newblock In \emph{Proc. of COLING'96}, pages 1041--1044.
\newblock \url{http://aclweb.org/anthology/C/C96/C96-2183.pdf}.

\bibitem[{Choshen and Abend(2018)}]{CA18}
Leshem Choshen and Omri Abend. 2018.
\newblock Reference-less measure of faithfulness for grammatical error
  correction.
\newblock In \emph{Proc. of NAACL'18 (Short papers)}.
\newblock To appear.

\bibitem[{Cohen(1968)}]{C68}
Jacob Cohen. 1968.
\newblock Weighted kappa: {N}ominal scale agreement provision for scaled
  disagreement or partial credit.
\newblock \emph{Psychological bulletin}, 70(4):213.

\bibitem[{Coster and Kauchak(2011{\natexlab{a}})}]{CK11b}
William Coster and David Kauchak. 2011{\natexlab{a}}.
\newblock {Learning to
  simplify sentences using {W}ikipedia}.
\newblock In \emph{Proc. of ACL, Short Papers}, pages 1--9.
\newblock \url{http://aclweb.org/anthology/P/P11/P11-2117.pdf}.

\bibitem[{Coster and Kauchak(2011{\natexlab{b}})}]{CK11a}
William Coster and David Kauchak. 2011{\natexlab{b}}.
\newblock {Simple
  {E}nglish {W}ikipedia: {A} new text simplification task}.
\newblock In \emph{Proc. of ACL'11}, pages 665--669.
\newblock \url{http://aclweb.org/anthology/P/P11/P11-2117.pdf}.

\bibitem[{Dixon(2010{\natexlab{a}})}]{D10B}
Robert~M.W. Dixon. 2010{\natexlab{a}}.
\newblock \emph{{B}asic {L}inguistic {T}heory: {G}rammatical {T}opics},
  volume~2.
\newblock Oxford University Press.

\bibitem[{Dixon(2010{\natexlab{b}})}]{D10A}
Robert~M.W. Dixon. 2010{\natexlab{b}}.
\newblock \emph{{B}asic {L}inguistic {T}heory: {M}ethodology}, volume~1.
\newblock Oxford University Press.

\bibitem[{Dixon(2012)}]{D12}
Robert~M.W. Dixon. 2012.
\newblock \emph{{B}asic {L}inguistic {T}heory: {F}urther {G}rammatical
  {T}opics}, volume~3.
\newblock Oxford University Press.

\bibitem[{Ganitketitch et~al.(2013)Ganitketitch, Durme, and
  Callison-Burch}]{G13}
Jury Ganitketitch, Benjamin~Van Durme, and Chris Callison-Burch. 2013.
\newblock {{PPDB}: {T}he
  paraphrase database}.
\newblock In \emph{Proc. of NAACL-HLT'13}, pages 758--764.
\newblock \url{http://www.aclweb.org/anthology/N13-1092}. 

\bibitem[{Gardent et~al.(2017)Gardent, Shimorina, Narayan, and
  Perez-Beltrachini}]{G17}
Claire Gardent, Anastasia Shimorina, Shashi Narayan, and Laura
  Perez-Beltrachini. 2017.
\newblock {Creating
  training corpora for {NLG} micro-planning}.
\newblock In \emph{Proc. of ACL'17}, pages 179--188.
\newblock \url{http://aclweb.org/anthology/P/P17/P17-1017.pdf}.

\bibitem[{Glava\v{s} and \v{S}tajner(2013)}]{GS13}
Goran Glava\v{s} and Sanja \v{S}tajner. 2013.
\newblock {Event-centered
  simplification of news stories}.
\newblock In \emph{Proc. of the Student Research Workshop associated with RANLP
  2013}, pages 71--78.
\newblock \url{http://www.aclweb.org/anthology/R13-2011}.

\bibitem[{Hershcovich et~al.(2017)Hershcovich, Abend, and Rappoport}]{H17}
Daniel Hershcovich, Omri Abend, and Ari Rappoport. 2017.
\newblock {A
  transition-based directed acyclic graph parser for {UCCA}}.
\newblock In \emph{Proc. of ACL'17}, pages 1127--1138.
\newblock \url{http://aclweb.org/anthology/P/P17/P17-1104.pdf}.

\bibitem[{Hung et~al.(2012)Hung, Minh, and Shimazu}]{H12}
Bui~Thanh Hung, Nguyen~Le Minh, and Akira Shimazu. 2012.
\newblock Sentence splitting for {V}ietnamese-{E}nglish machine translation.
\newblock In \emph{Knowledge and Systems Engineering, 2012 Fourth International
  Conference}, pages 156--160.

\bibitem[{Hwang et~al.(2015)Hwang, Hajishirzi, Ostendorf, and Wu}]{H15}
William Hwang, Hannaneh Hajishirzi, Mari Ostendorf, and Wei Wu. 2015.
\newblock {Aligning
  sentences from {S}tandard {W}ikipedia to {S}imple {W}ikipedia}.
\newblock In \emph{Proc. of NAACL'15}, pages 211--217.
\newblock \url{http://aclweb.org/anthology/N/N15/N15-1022.pdf}.

\bibitem[{Kamp(1981)}]{K81}
Hans Kamp. 1981.
\newblock A theory of truth and semantic representation.
\newblock In \emph{Formal methods in the study of language}. Mathematisch
  Centrum.
\newblock Number pt.1 in Mathematical Centre tracts.

\bibitem[{Kauchak(2013)}]{K13}
David Kauchak. 2013.
\newblock {Improving
  text simplification language modeling using unsimplified text data}.
\newblock In \emph{Proc. of ACL'13}, pages 1537--1546.
\newblock \url{http://aclweb.org/anthology/P/P13/P13-1151.pdf}

\bibitem[{Klein et~al.(2017)Klein, Kim, Deng, Senellart, and Rush}]{K17}
Guillam Klein, Yoon Kim, Yuntian Deng, Jean Senellart, and Alexander~M. Rush.
  2017.
\newblock {Open {NMT}:
  {O}pen-source toolkit for neural machine translation}.
\newblock ArXiv:1701.02810 [cs:CL].
\newblock \url{https://arxiv.org/pdf/1701.02810.pdf}.

\bibitem[{Koehn et~al.(2007)Koehn, Hoang, Birch, Callison-Buch, Federico,
  Bertoldi, Cowan, Shen, Moran, Zens, Dyer, Bojar, Constantin, and
  Herbst}]{K07}
Philipp Koehn, Hieu Hoang, Alexandra Birch, Chris Callison-Buch, Marcello
  Federico, Nicola Bertoldi, Brooke Cowan, Wade Shen, Christine Moran, Richard
  Zens, Chris Dyer, Ond\v{r}ej Bojar, Alexandra Constantin, and Evan Herbst.
  2007.
\newblock {Moses: open
  source toolkit for statistical machine translation}.
\newblock In \emph{Proc. of ACL'07 on interactive poster and demonstration
  sessions}, pages 177--180.
\newblock \url{http://aclweb.org/anthology/P/P07/P07-2045.pdf}.

\bibitem[{Koehn et~al.(2003)Koehn, Och, and Marcu}]{K03}
Philipp Koehn, Franz~Josef Och, and Daniel Marcu. 2003.
\newblock {Statistical
  phrase-based translation}.
\newblock In \emph{Proc. of NAACL'03}, pages 48--54.
\newblock \url{http://aclweb.org/anthology/N/N03/N03-1017.pdf}.

\bibitem[{Langacker(2008)}]{L08}
Ronald~W. Langacker. 2008.
\newblock \emph{{C}ognitive {G}rammar: {A} {B}asic {I}ntroduction}.
\newblock Oxford University Press, USA.

\bibitem[{Loper and Bird(2002)}]{LB02}
Edward Loper and Steven Bird. 2002.
\newblock {{NLTK}:
  the natural language toolkit}.
\newblock In \emph{Proc. of EMNLP'02}, pages 63--70.
\newblock \url{http://www.aclweb.org/anthology/W/W02/W02-0109.pdf}.

\bibitem[{Luong et~al.(2015)Luong, Pham, and Manning}]{L15}
Minh-Thang Luong, Hieu Pham, and Christopher~D. Manning. 2015.
\newblock {Effective approaches to
  attention-based neural machine translation}.
\newblock In \emph{Proc. of EMNLP'15}, pages 1412--1421.
\newblock \url{http://aclweb.org/anthology/D15-1166}.

\bibitem[{Mendes et~al.(2012)Mendes, Jakob, and Bizer}]{M12}
Pablo~N Mendes, Max Jakob, and Christian Bizer. 2012.
\newblock {{DB}pedia:
  {A} multilingual cross-domain knowledge base}.
  \newblock In \emph{Proc. of LREC'12}, pages 1813--1817.
  \newblock \url{http://www.lrec-conf.org/proceedings/lrec2012/pdf/570_Paper.pdf}.

\bibitem[{Mikolov et~al.(2013{\natexlab{a}})Mikolov, Chen, Corrado, and
  Dean}]{M13a}
Tomas Mikolov, Kai Chen, Greg Corrado, and Jeffrey Dean. 2013{\natexlab{a}}.
\newblock Efficient estimation of word representations in vector space.
\newblock In \emph{Proc. of Workshop at International Conference on Learning
  Representations}.

\bibitem[{Mikolov et~al.(2013{\natexlab{b}})Mikolov, Sutskever, Chen, Corrado,
  and Dean}]{M13b}
Tomas Mikolov, Ilya Sutskever, Kai Chen, Greg~S Corrado, and Jeff Dean.
  2013{\natexlab{b}}.
\newblock 
  {Distributed representations of words and phrases and their
  compositionality}.
\newblock In \emph{Advances in neural information processing systems}, pages
3111--3119.
\newblock \url{https://papers.nips.cc/paper/5021-distributed-representations-} \url{of-words-and-phrases-and-their-} \url{compositionality.pdf}.

\bibitem[{Mishra et~al.(2014)Mishra, Soni, Sharma, and Sharma}]{M14}
Kshitij Mishra, Ankush Soni, Rahul Sharma, and Dipti~Misra Sharma. 2014.
\newblock {Exploring the
  effects of sentence simplification on {H}indi to {E}nglish {M}achine
  {T}ranslation systems}.
\newblock In \emph{Proc. of the Workshop on Automatic Text Simplification:
  Methods and Applications in the Multilingual Society}, pages 21--29.
\newblock \url{http://www.aclweb.org/anthology/W14-5603}.

\bibitem[{Narayan and Gardent(2014)}]{NG14}
Shashi Narayan and Claire Gardent. 2014.
\newblock {Hybrid
  simplification using deep semantics and machine translation}.
\newblock In \emph{Proc. of ACL'14}, pages 435--445.
\newblock \url{http://aclweb.org/anthology/P/P14/P14-1041.pdf}.

\bibitem[{Narayan and Gardent(2016)}]{NG16}
Shashi Narayan and Claire Gardent. 2016.
\newblock {Unsupervised
  sentence simplification using deep semantics}.
\newblock In \emph{Proc. of INLG'16}, pages 111--120.
\newblock \url{http://aclweb.org/anthology/W/W16/W16-6620.pdf}.

\bibitem[{Narayan et~al.(2017)Narayan, Gardent, Cohen, and Shimorina}]{N17}
Shashi Narayan, Claire Gardent, Shay~B. Cohen, and Anastasia Shimorina. 2017.
\newblock {Split and
  rephrase}.
\newblock In \emph{Proc. of EMNLP'17}, pages 617--627.
\newblock \url{http://aclweb.org/anthology/D/D17/D17-1064.pdf}.

\bibitem[{Niklaus et~al.(2016)Niklaus, Bermeitinger, Handschuh, and
  Freitas}]{N16}
Christina Niklaus, Bernahard Bermeitinger, Siegfried Handschuh, and Andr\'{e}
  Freitas. 2016.
\newblock {A sentence
  simplification system for improving relation extraction}.
\newblock In \emph{Proc. of COLING'16}.
\newblock \url{http://aclweb.org/anthology/C/C16/C16-2036.pdf}.

\bibitem[{Nisioi et~al.(2017)Nisioi, \v{S}tajner, Ponzetto, and Dinu}]{Ni17}
Sergiu Nisioi, Sanja \v{S}tajner, Simone~Paolo Ponzetto, and Liviu~P. Dinu.
  2017.
\newblock {Exploring neural
  text simplification models}.
\newblock In \emph{Proc. of ACL'17 (Short paper)}, pages 85--91.
\newblock \url{http://www.aclweb.org/anthology/P17-2014}.

\bibitem[{Papineni et~al.(2002)Papineni, Roukos, Ward, and Zhu}]{P02}
Kishore Papineni, Salim Roukos, Todd Ward, and Wei-Jing Zhu. 2002.
\newblock {{BLEU}: a
  method for automatic evaluation of machine translation}.
\newblock In \emph{Proc. of ACL'02}, pages 311--318.
\newblock \url{http://aclweb.org/anthology/P/P02/P02-1040.pdf}.

\bibitem[{{\v R}eh\r{u}\v{r}ek and Sojka(2010)}]{RS10}
Radim {\v R}eh\r{u}\v{r}ek and Petr Sojka. 2010.
\newblock Software framework for topic modelling with large corpora.
\newblock In \emph{Proc. of LREC 2010 Workshop on New Challenges for NLP
  Frameworks}, pages 45--50, Valletta, Malta. ELRA.

\bibitem[{Rello et~al.(2013)Rello, Baeza-Yates, Bott, and Saggion}]{R13}
Luz Rello, Ricardo Baeza-Yates, Stefan Bott, and Horacio Saggion. 2013.
\newblock Simplify or help?: text simplification strategies for people with
  dyslexia.
\newblock In \emph{Proc. of the 10th International Cross-Disciplinary
  Conference on Web Accesibility}, pages 15:1 -- 15:10.

\bibitem[{Seretan(2012)}]{S12}
Violeta Seretan. 2012.
\newblock 
  {Acquisition of syntactic simplification rules for {F}rench}.
\newblock In \emph{Proc. of LREC'12}, pages 4019--4026.
\newblock \url{ http://www.lrec-conf.org/proceedings/lrec2012/pdf/304_Paper.pdf}. 

\bibitem[{Shardlow(2014)}]{Sh14}
Matthew Shardlow. 2014.
\newblock {A survey
  of automated text simplification}.
\newblock \emph{International Journal of Advanced Computer Science and
  Applications}.
\newblock \url{https://doi.org/10.14569/SpecialIssue.2014.040109}.

\bibitem[{Siddharthan(2002)}]{S02}
Advaith Siddharthan. 2002.
\newblock An architecture for a text simplification system.
\newblock In \emph{Proc. of LEC}, pages 64--71.

\bibitem[{Siddharthan(2004)}]{S04}
Advaith Siddharthan. 2004.
\newblock Syntactic simplification and text cohesion.
\newblock Technical Report 597, University of Cambridge.

\bibitem[{Siddharthan and Angrosh(2014)}]{SA14}
Advaith Siddharthan and M.~A. Angrosh. 2014.
\newblock {Hybrid text
  simplification using synchronous dependency grammars with hand-written and
  automatically harvested rules}.
\newblock In \emph{Proc. of EACL'14}, pages 722--731.
\newblock \url{http://aclweb.org/anthology/E/E14/E14-1076.pdf}.

\bibitem[{Siddhathan(2011)}]{S11}
Advaith Siddhathan. 2011.
\newblock {Text
  simplification using typed dependencies: {A} comparison of the robustness of
  different generation strategies}.
\newblock In \emph{Proc. of the 13th European Workshop on Natural Language
  Generation}, pages 2--11. Association of Computational Linguistics.
\newblock \url{http://aclweb.org/anthology/W/W11/W11-2802.pdf}.

\bibitem[{Smith and Eisner(2006)}]{SE06}
David~A. Smith and Jason Eisner. 2006.
\newblock 
  {Quasi-synchronous grammars: {A}lignment by soft projection of syntactic
  dependencies}.
\newblock In \emph{Proc. of the 1st Workshop in Statistical Machine
  Translation}, pages 23--30.
\newblock \url{http://aclweb.org/anthology/W/W06/W06-3104.pdf}.

\bibitem[{Specia(2010)}]{S10}
Lucia Specia. 2010.
\newblock Translating from complex to simplified sentences.
\newblock In \emph{Proc. of the 9th International Conference on Computational
  Processing of the Portuguese Language}, pages 30--39.

\bibitem[{Srivastava et~al.(2014)Srivastava, Hinton, Krizhevsky, Sutskever, and
  Salakhutdinov}]{Sr14}
Nitish Srivastava, Geoffrey~E Hinton, Alex Krizhevsky, Ilya Sutskever, and
  Ruslan Salakhutdinov. 2014.
\newblock {Dropout: a
  simple way to prevent neural networks from overfitting}.
\newblock \emph{Journal of Machine Learning Research}, 15(1):1929--1958.
\newblock \url{http://jmlr.org/papers/volume15/srivastava14a/srivastava14a.pdf}. 

\bibitem[{{\v S}tajner et~al.(2015){\v S}tajner, Bechara, and Saggion}]{Sa15}
Sanja {\v S}tajner, Hannah Bechara, and Horacio Saggion. 2015.
\newblock {A deeper
  exploration of the standard {PB-SMT} approach to text simplification and its
  evaluation}.
\newblock In \emph{Proc. of ACL'15 (Short papers)}, pages 823--828.
\newblock \url{http://aclweb.org/anthology/P/P15/P15-2135.pdf}.

\bibitem[{{\v S}tajner and Glava\v{s}(2017)}]{SG17}
Sanja {\v S}tajner and Goran Glava\v{s}. 2017.
\newblock Leveraging event-based semantics for automated text simplification.
\newblock \emph{Expert systems with applications}, 82:383--395.

\bibitem[{{\v S}tajner and Popovi\'{c}(2016)}]{SP16}
Sanja {\v S}tajner and Maja Popovi\'{c}. 2016.
\newblock Can text simplification help machine translation.
\newblock \emph{Baltic J. Modern Computing}, 4:230--242.

\bibitem[{Sulem et~al.(2015)Sulem, Abend, and Rappoport}]{S15}
Elior Sulem, Omri Abend, and Ari Rappoport. 2015.
\newblock {Conceptual
  annotations preserve structure across translations}.
\newblock In \emph{Proc. of 1st Workshop on Semantics-Driven Statistical
  Machine Translation (S2Mt 2015)}, pages 11--22.
\newblock \url{http://aclweb.org/anthology/W/W15/W15-3502.pdf}.

\bibitem[{Sulem et~al.(2018)Sulem, Abend, and Rappoport}]{S18}
Elior Sulem, Omri Abend, and Ari Rappoport. 2018.
\newblock Semantic structural evaluation for text simplification.
\newblock In \emph{Proc. of NAACL'18}.
\newblock To appear.

\bibitem[{Woodsend and Lapata(2011)}]{WL11}
Kristian Woodsend and Mirella Lapata. 2011.
\newblock {Learning to
  simplify sentences with quasi-synchronous grammar and integer programming}.
\newblock In \emph{Proc. of EMNLP'11}, pages 409--420.
\newblock \url{http://aclweb.org/anthology/D/D11/D11-1038.pdf}.

\bibitem[{Wubben et~al.(2012)Wubben, van~den Bosch, and Krahmer}]{W12}
Sander Wubben, Antal van~den Bosch, and Emiel Krahmer. 2012.
\newblock {Sentence
  simplification by monolingual machine translation}.
\newblock In \emph{Proc. of ACL'12}, pages 1015--1024.
\newblock \url{http://aclweb.org/anthology/P/P12/P12-1107.pdf}.

\bibitem[{Xu et~al.(2015)Xu, Callison-Burch, and Napoles}]{Xu15}
Wei Xu, Chris Callison-Burch, and Courtney Napoles. 2015.
\newblock {Problems in
  current text simplification research: new data can help}.
\newblock \emph{TACL}, 3:283--297.
\newblock \url{http://aclweb.org/anthology/Q/Q15/Q15-1021.pdf}.

\bibitem[{Xu et~al.(2016)Xu, Napoles, Pavlick, Chen, and Callison-Burch}]{Xu16}
Wei Xu, Courtney Napoles, Ellie Pavlick, Quanze Chen, and Chris Callison-Burch.
  2016.
\newblock {Optimizing
  statistical machine translation for text simplification}.
\newblock \emph{TACL}, 4:401--415.
\newblock \url{http://aclweb.org/anthology/Q/Q16/Q16-1029.pdf}.

\bibitem[{Yamada and Knight(2001)}]{YK01}
Kenji Yamada and Kevin Knight. 2001.
\newblock {A syntax-based
  statistical translation model}.
\newblock In \emph{Proc.of ACL'01}, pages 523--530.
\newblock \url{http://www.aclweb.org/anthology/P01-1067}.

\bibitem[{Zhang and Lapata(2017)}]{ZL17}
Xingxing Zhang and Mirella Lapata. 2017.
\newblock {Sentence simplification
  with deep reinforcement learning}.
\newblock In \emph{Proc. of EMNLP'17}, pages 595--605.
\newblock \url{http://aclweb.org/anthology/D17-1062}.

\bibitem[{Zhang et~al.(2017)Zhang, Ye, Zhao, and Yan}]{Z17}
Yaoyuan Zhang, Zhenxu Ye, Dongyan Zhao, and Rui Yan. 2017.
\newblock {A constrained
  sequence-to-sequence neural model for sentence simplification}.
\newblock ArXiv:1704.02312 [cs.CL].
\newblock \url{https://arxiv.org/pdf/1704.02312.pdf}.

\bibitem[{Zhu et~al.(2010)Zhu, Bernhard, and Gurevych}]{Z10}
Zhemin Zhu, Delphine Bernhard, and Iryna Gurevych. 2010.
\newblock {A monolingual
  tree-based translation model for sentence simplification}.
\newblock In \emph{Proc. of COLING'10}, pages 1353--1361.
\newblock \url{http://aclweb.org/anthology/C/C10/C10-1152.pdf}.

\end{thebibliography}
\bibliographystyle{acl_natbib}

\end{document}